\documentclass[graybox]{svmult}

\usepackage{mathptmx}       
\usepackage{helvet}         
\usepackage{courier}        
\usepackage{textcomp}
\usepackage{makeidx}         
\usepackage{graphicx}        
\usepackage{multicol}        
\usepackage[bottom]{footmisc}

\usepackage[ruled,vlined,linesnumbered]{algorithm2e} 
\usepackage{listings} 
\usepackage{xcolor}
\usepackage{amsmath}
\usepackage[caption=false]{subfig}

\definecolor{mutedBlue}{HTML}{4C72B0}
\definecolor{mutedGreen}{HTML}{677D29}
\usepackage{tikz}
\usetikzlibrary{matrix,backgrounds}
\usepackage{tikz-qtree}

\makeindex             

\begin{document}

\title*{Symbolic Regression by Exhaustive Search -- Reducing the Search Space using Syntactical Constraints and Efficient Semantic Structure Deduplication.}
\titlerunning{Symbolic Regression by Exhaustive Search}
\author{Lukas Kammerer and Gabriel Kronberger and Bogdan Burlacu and Stephan M. Winkler and Michael Kommenda and Michael Affenzeller}
\authorrunning{Kammerer et al.}
\institute {
 Lukas Kammerer$^{1,2,3}$ \email{lukas.kammerer@fh-hagenberg.at}
\and Gabriel Kronberger$^{1,3}$
\and Bogdan Burlacu$^{1,3}$
\and Stephan M. Winkler$^{1,2}$
\and Michael Kommenda$^{1,3}$
\and Michael Affenzeller$^{1,2}$
    \at
    \inst{1} Heuristic and Evolutionary Algorithms Laboratory (HEAL),
    University of Applied Sciences Upper Austria, Softwarepark 11, 4232 Hagenberg, Austria\\
    \inst{2} Department of Computer Science,
    Johannes Kepler University, Altenberger Stra{\ss}e 69, 4040 Linz, Austria \\
    \inst{3} Josef Ressel Center for Symbolic Regression,
    University of Applied Sciences Upper Austria, Softwarepark 11, 4232 Hagenberg, Austria\\
}
%
%
\maketitle

\renewcommand{\thefootnote}{}
\footnotetext{\hspace{-0em}
	The final publication is available at \url{https://link.springer.com/chapter/10.1007\%2F978-3-030-39958-0\_5}.
}
\renewcommand\thefootnote{\arabic{footnote}}

\abstract*{Symbolic regression is a powerful system identification technique in industrial scenarios where no prior knowledge on model structure is available. Such scenarios often require specific model properties such as interpretability, robustness, trustworthiness and plausibility, that are not easily achievable using standard approaches like genetic programming for symbolic regression. In this chapter we introduce a deterministic symbolic regression algorithm specifically designed to address these issues.
The algorithm uses a context-free grammar to produce models that are parameterized by a non-linear least squares local optimization procedure. A finite enumeration of all possible models is guaranteed by structural restrictions as well as a caching mechanism for detecting semantically equivalent solutions.
Enumeration order is established via heuristics designed to improve search efficiency. 
Empirical tests on a comprehensive benchmark suite show that our approach is competitive with genetic programming in many noiseless problems while maintaining desirable properties such as simple, reliable models and reproducibility.    
}
\abstract{Symbolic regression is a powerful system identification technique in industrial scenarios where no prior knowledge on model structure is available. Such scenarios often require specific model properties such as interpretability, robustness, trustworthiness and plausibility, that are not easily achievable using standard approaches like genetic programming for symbolic regression. In this chapter we introduce a deterministic symbolic regression algorithm specifically designed to address these issues.
The algorithm uses a context-free grammar to produce models that are parameterized by a non-linear least squares local optimization procedure. A finite enumeration of all possible models is guaranteed by structural restrictions as well as a caching mechanism for detecting semantically equivalent solutions.
Enumeration order is established via heuristics designed to improve search efficiency. 
Empirical tests on a comprehensive benchmark suite show that our approach is competitive with genetic programming in many noiseless problems while maintaining desirable properties such as simple, reliable models and reproducibility.    
}


\section{Introduction}

Symbolic regression is a task that we can solve with
genetic programming (GP) and a common example where GP is
particularly effective in practical applications. Symbolic
regression is a machine learning task whereby we try to find a
mathematical model represented as a closed-form expression
that captures dependencies of variables from a dataset.
Genetic programming has been proven to be well-suited for
this task especially when there is little knowledge about the
data-generating process. Even when we have a good understanding of the
underlying process, GP can identify counterintuitive or unexpected
solutions.

\subsection{Motivation}

GP has some practical limitations when used for symbolic regression. One limitation
is that---as a stochastic process---it might produce highly dissimilar
solutions even for the same input data. This can be very helpful to
produce new ``creative'' solutions. However, it is problematic when we try
to integrate symbolic regression in carefully engineered solutions
(e.g. for automatic control of production plants). In such situations
we would hope that there is an optimal solution and the solution
method guarantees to identify the optimum. Intuitively, if the data
changes only slightly, we expect that the optimal regression solution
also changes only slightly. If this is the case we know that the
solution method is trustworthy (cf.~\cite{Kotanchek:2007:GPTP,Stijven:2015:GPTP}) and we
can rely on the fact that the solutions are optimal at least with
respect to the objective function that we specified. Of course this is
only wishful thinking because of three fundamental reasons: (1) the
symbolic regression search space is huge and contains many different
expressions which are algebraically equivalent, (2) GP has no
guarantee to explore the whole search space with reasonable
computational resources and (3) the "optimal solution" might not 
be expressible as a closed-form mathematical expressions using 
the given building blocks. Therefore, the goal is to find 
an approximately optimal solution.

\subsection{Prior Work}

Different methods have been developed with
the aim to improve the reliability of symbolic regression. Currently,
there are several off-the-shelf software solutions which use
enhanced variants of GP and are noteworthy in this context: the
DataModeler
package\footnote{\url{http://www.evolved-analytics.com/}} \cite{Kotanchek2013}
provides extensive capabilities for symbolic regression on top of
Mathematica\texttrademark. Eureqa\texttrademark\, is a commercial
software
tool\footnote{\url{https://www.nutonian.com/products/eureqa/}} for
symbolic regression based on research described in
\cite{Schmidt:2006:GPTP,Schmidt:2009:GPTP,Schmidt:2010:GPTP}. The open-source framework HeuristicLab\footnote{\url{https://dev.heuristiclab.com}} \cite{wagner2005heuristiclab} is a general software environment for heuristic and
evolutionary algorithms with extensive functionality for symbolic regression and white-box modeling.

In other prior work, several researchers have presented
non-evolutionary solution methods for symbolic regression. Fast
function extraction (FFX) \cite{McConaghy:2011:GPTP} is a deterministic
method that uses elastic-net regression \cite{zou2005regularization}
to produce symbolic
regression solutions orders of magnitudes faster than GP for many
real-world problems.
The work by Korns toward ``extremely accurate'' symbolic regression
\cite{Korns:2013:GPTP,Korns:2014:GPTP,Korns:2015:GPTP} highlights the
issue that baseline GP does not guarantee to find the optimal solution
even for rather limited search spaces. They give a useful systematic
definition of increasingly larger symbolic regression search spaces
using abstract expression grammars \cite{Korns:2009:GEC} and describes
enhancements to GP to improve it's reliability. The work by Worm and
Chiu on prioritized grammar enumeration \cite{Worm:2013:GECCO} is
closely related. They use a restricted set of grammar rules for
deriving increasingly complex expressions and describe a deterministic
search algorithm, which enumerates the
search space for limited symbolic regression problems.

\subsection{Organization of this Chapter}

Our contribution is conceptually an extension of prioritized grammar enumeration \cite{Worm:2013:GECCO}, although our implementation of the method deviates significantly. The most relevant extensions are that we cut out large parts of the search space and provide a general framework for integrating heuristics in order to improve the search efficiency. Section \ref{sec:search_space} describes how we reduce the size of the search space which is defined by a context-free grammar:

\begin{enumerate}
    \item We restrict the structure of solution to prevent too complicated solutions.
    \item We use grammar restrictions to prevent semantic duplicates---solutions with different syntax but same semantics, such as algebraic transformations. With these restrictions, most solutions can only be generated in exactly one way.
    \item We efficiently identify remaining duplicates with semantic hashing, so that (nearly) all solutions in the search space are semantically unique.
\end{enumerate}

In Section \ref{sec:exploring_search_space}, we explain the algorithm that iterates all these semantically unique solutions. The algorithm sequentially generates solutions from the grammar and keeps track of the most accurate one. For very small problems, it is even feasible to iterate the whole search space \cite{Kronberger2019}. However, our goal in larger problems is to find accurate and concise solutions early during the search and to stop the algorithm after a reasonable time. The search order is determined with heuristics, which estimate the quality of solutions and prioritize promising ones in the search. A simple heuristic is proposed in Section \ref{sec:search_heuristic}. Modeling results in Section \ref{sec:experiments} show that this first version of our algorithm can already solve several difficult noiseless benchmark problems.

\section{Definition of the Search Space}
\label{sec:search_space}

The search space of our deterministic symbolic regression algorithm is defined by a context-free grammar. Production rules in the grammar define the mathematical expressions that can be explored by the algorithm. The grammar only specifies possible model structures whereby placeholders are used for numeric coefficients. These are optimized separately by a curve-fitting algorithm (e.g.~optimizing least squares with an gradient-based optimization algorithm) using the available training data.

In a general grammar for mathematical expressions---as it is common in symbolic regression with GP for example---the same formula can be derived in several forms. These duplicates inflate the search space. To reduce them, our grammar is deliberately restricted regarding the possible structure of expressions. Remaining duplicates that cannot be prevented by a context-free grammar are eliminated via a hashing algorithm. Using both this grammar and hashing, we can generate a search space with only semantically unique expressions.

\subsection{Grammar for Mathematical Expressions}

In this work we consider mathematical expressions as list of symbols which we call \textit{phrases} or \textit{sentences}. A phrase can contain both \textit{terminal} and \textit{non-terminal} symbols and a sentence only terminal symbols. Non-terminal symbols can be replaced by other symbols as defined by a grammar's \textit{production rules} while terminal symbols represent parts of the final expression like functions or variables in our case.

Our grammar is very similar to the one by Kronberger et al.~\cite{Kronberger2019}. It produces only rational polynomials which may contain linear and nonlinear terms, as outlined conceptually in Equation~\ref{eq:polynomial_structure}. The basic building blocks of terms are linear and non-linear functions $\{+, \times$, inv, exp, log, sin, square root, cube root$\}$. Recursion in the production rules represents a strategy for generating increasingly complex solutions by repeated nesting of expressions and terms. 

\begin{equation}
    \begin{split}
    \mathit{Expr} &= c_1 \mathit{Term}_1 + c_2 \mathit{Term}_2 + \ldots + c_n\\
    \mathit{Term} &= \mathit{Factor}_0 \times \mathit{Factor}_1 \times \ldots \\
    \mathit{Factor} &\in \{\mathit{variable}, \log(\mathit{\mathit{variable}}), \exp(\mathit{\mathit{variable}}), \sin(\mathit{\mathit{variable}}) \} \\
    \end{split}
    \label{eq:polynomial_structure}
\end{equation}

We explicitly disallow nested non-linear functions, as we consider such solutions too complex for real-world applications. Otherwise, we allow as many different structures as possible to keep accurate and concise models in the search space. We prevent semantic duplicates by generating just one side of mathematical equality relations in our grammar, e.g.~we allow $xy+xz$ but not $x(y+z)$. Since each function has different mathematical identities, many different production rules are necessary to cover all special cases. Because we scale every term including function arguments, we also end up with many placeholders for coefficients in the structures. All production rules are detailed in Listing~\ref{lst:grammar} and described in the following. 

\begin{lstlisting}[caption={Context-free grammar for generating mathematical expressions}, label={lst:grammar}, language=C]
G(Expr):
// Expressions and terms for polynomial structure
    Expr       -> "const" "*" Term "+" Expr    | 
                  "const" "*" Term "+" "const"
 
    Term       -> RecurringFactors "*" Term    | 
                  RecurringFactors             |
                  OneTimeFactors
        
    RecurringFactors -> VarFactor | LogFactor |
                        ExpFactor | SinFactor
 
    VarFactor  -> <variable>
    LogFactor  -> "log" "(" SimpleExpr ")" 
    ExpFactor  -> "exp" "(" "const" "*" SimpleTerm ")"
    SinFactor  -> "sin" "(" SimpleExpr ")"
    
// Factors which can occur at most once per term
    OneTimeFactors -> InvFactor "*" SqrtFactor "*" CbrtFactor |
                      InvFactor "*" SqrtFactor                |
                      InvFactor "*"                CbrtFactor |
                                    SqrtFactor "*" CbrtFactor |
                      InvFactor                               |
                                    SqrtFactor                |
                                                   CbrtFactor

    InvFactor  -> "1/" "(" InvExpr ")"
    SqrtFactor -> "sqrt" "(" SimpleExpr ")"
    CbrtFactor -> "cbrt" "(" SimpleExpr ")"
    
// Function arguments
    SimpleExpr -> "const" "*" SimpleTerm "+" SimpleExpr  | 
                  "const" "*" SimpleTerm "+" "const"
 
    SimpleTerm -> VarFactor "*" SimpleTerm | VarFactor 
    
    InvExpr -> "const" "*" InvTerm "+" InvExpr |
               "const" "*" InvTerm "+" "const"
 
    InvTerm -> RecurringFactors "*" InvTerm  | 
               RecurringFactors "*" SqrtFactor "*" CbrtFactor |
               RecurringFactors "*" SqrtFactor                |
               RecurringFactors "*"                CbrtFactor |
                                    SqrtFactor "*" CbrtFactor |
               RecurringFactors                               |
                                    SqrtFactor                |
                                                   CbrtFactor
\end{lstlisting}


We use a polynomial structure as outlined in Equation~\ref{eq:polynomial_structure} to prevent a factored form of solutions. The polynomial structure is enforced with the production rules \texttt{Expr} and \texttt{Term}. We restrict the occurrence of the multiplicative inverse ($= \frac{1}{\ldots}$), the square root and cube root function to prevent a factored form such as $\frac{1}{x+y} \frac{1}{x+z}$. This is necessary since we want to allow sums of simple terms as function arguments (see non-terminal symbol \texttt{SimpleExpr}). Therefore, these three functions can occur at most once time per term. This is defined with symbol \texttt{OneTimeFactors} and one production rule for each combination. The only function in which we do not allow sums as argument is exponentiation (see \texttt{ExpFactor}), since this form is substituted by the overall polynomial structure (e.g.~we allow $e^x e^y$ but not $e^{x+y}$). Equation \ref{eq:covered_identities} shows some example identities and which forms are supported.

\begin{equation}
\begin{split}
\text{in the search space:} \quad \quad \quad \hphantom{\equiv} & \quad \quad \quad \text{not in the search space:} \\[3pt]
c_1 xy + c_2 xz + c_3                                
    \ \ \equiv & \ \
    x(c_4 y + c_5 z) + c_6 \\[3pt]
c_1 \frac{1}{c_2 x + c_3 xx + c_4 xy + c_5 y + c_6} + c_7 
    \ \ \equiv & \ \
    c_8 \frac{1}{c_{9} x + c_{10}} \ \frac{1}{c_{11} x + c_{12} y + c_{13}} + c_{14} \\[3pt]
c_1 \exp(c_2 x) \exp(c_3 y) + c_4
    \ \ \equiv & \ \
    c_5 \exp(c_6 x + c_7 y) + c_8
\end{split}
\label{eq:covered_identities}
\end{equation}

We only allow (sums of) terms of variables as function arguments, which we express with the production rules \texttt{SimpleExpr} and \texttt{SimpleTerm}. An exception is the multiplicative inverse, in which we want to include the same structures as in ordinary terms. However, we disallow compound fractions like in Equation \ref{eq:covered_identities_inverse}. Again, we introduce separate grammar rules \texttt{InvExpr} and \texttt{InvTerm} which cover the same rules as \texttt{Term} except the multiplicative inverse.

\begin{equation}
    \begin{split}
    \text{in the search space:} \quad \quad \quad \hphantom{\equiv} & \quad \quad \quad \text{not in the search space:} \\[3pt]
    c_1 \frac{1}{c_2 \log(c_3 x + c_4) + c_5} + c_6 
    \ \ \equiv & \ \
    c_7 \frac{1}{c_8 \frac{1}{c_9 \log(c_{10} x + c_{11}) + c_{12}} + c_{13}} + c_{14}
    \end{split}
    \label{eq:covered_identities_inverse}
\end{equation}

In the simplest case, the grammar produces an expression $E_0 = c_0 x+c_1$, where $x$ is a variable and $c_0$ and $c_1$ are coefficients corresponding to the slope and intercept. This expression is obtained by considering the simplest possible \texttt{Term} which corresponds to the derivation chain \texttt{Expr} $\to$ \texttt{Term} $\to$ \texttt{RecurringFactors} $\to$ \texttt{VarFactor} $\to$ $x$. Further derivations could lead for example to the expression $E_1 = c_0 x + (c_1 x + c_2)$, produced by nesting $E_0$ into the first part of the production rule for \texttt{Expr}, where the \texttt{Term} is again substituted with the variable $x$.


However, duplicate derivations can still occur due to algebraic properties like associativity and commutativity. These issues cannot be prevented with a context-free grammar because a context-free grammar does not consider surrounding symbols of the derived non-terminal symbol in its production rules. For example the expression $E_1 = c_0 x + (c_1 x + c_2)$ contains two coefficients $c_0$ and $c_1$ for variable $x$ which could be folded into a new coefficient $c_\mathit{new} = c_0 + c_1$. This type of redundancy becomes even more pronounced when \texttt{VarFactor} has multiple productions (corresponding to multiple input variables), as it becomes possible for multiple derivation paths to produce different expressions which are algebraically equivalent, such as $c_1x + c_2y$, $c_3x+ c_4x + c_5y$, $c_6y+c_7x$ for corresponding values of $c_1 ... c_7$. Another example are $c_1 x y$ and $c_2 y x$ which are both equivalent but derivable from the grammar. 

To avoid re-visiting already explored regions of the search space, we implement a caching strategy based on expression hashing for detecting algebraically equivalent expressions. The computed hash values are the same for algebraically equivalent expressions. In the search algorithm we keep the hash values of all visited expressions and prevent re-evaluations of expressions with identical hash values.

\subsection{Expression Hashing}

We employ expression hashing by Burlacu et al.~\cite{Burlacu2019eurocast} to assign hash values to subexpressions within phrases and sentences. Hash values for parent expressions are aggregated in a bottom-up manner from the hash values of their children using any general-purpose hash function. 
We then simplify such expressions according to arithmetic properties such as commutativity, associativity, and applicable mathematical identities. The resulting canonical minimal form and associated hash value are then cached in order to prevent duplicated search effort. 

\begin{figure}
    \centering
    \begin{tikzpicture}
    \tikzset{level distance=18mm, sibling distance=6mm}
    \Tree [.\node[draw,align=left]{$\mathbf{Root} = \text{``}+ \text{''}$\\$H=\oplus(\mathrm{Root},H_0,H_1)$ }; 
        [.\node[draw,align=left]{$\mathbf{N_0} = \text{``}\times\text{''}$\\$H_0 = \oplus(N_0, H_{0,0}, H_{0,1})$};  
            [.\node[draw,align=left]{$\mathbf{N_{0,0}} = c_1$\\$H_{0,0} = \oplus(N_{0,0})$};] 
            [.\node[draw,align=left]{$\mathbf{N_{0,1}} = x_1$\\$H_{0,1} = \oplus(N_{0,1})$};] 
        ]                   
        [.\node[draw,align=left]{$\mathbf{N_1} = \text{``}\times\text{''}$\\$H_1 = \oplus(N_1, H_{1,0}, H_{1,1})$};
            [.\node[draw,align=left]{$\mathbf{N_{1,0}} = c_2$\\$H_{1,0} = \oplus(N_{1,0})$};] 
            [.\node[draw,align=left]{$\mathbf{N_{1,1}} = x_2$\\$H_{1,1} = \oplus(N_{1,1})$};] 
        ]
    ]
\end{tikzpicture}
    \caption{Hash tree example, in which the hash values of all nodes are calculated from both their own node content and the has value of their children \cite{Burlacu2019eurocast}.}\label{fig:hash-tree}
\end{figure}
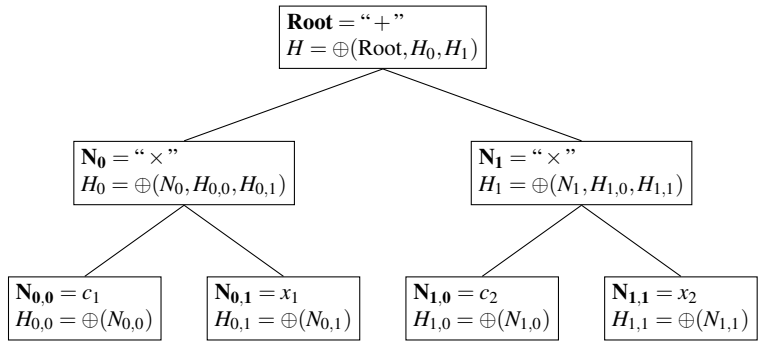

Expression hashing builds on the idea of Merkle trees \cite{Merkle1988}. Figure \ref{fig:hash-tree} shows how hash values propagate towards the tree root (the topmost symbol of the expression) using hash function $\oplus$ to aggregate child and parent hash values. Expression hashing considers an internal node's own symbol, as well as associativity and commutativity properties. To account for these properties, each hashing step must be accompanied by a corresponding sorting step, where child subexpressions are reordered according to their type and hash value. Algorithm~\ref{alg:expression-hashing} ensures that child nodes are sorted and hashed before parent nodes, such that calculated hash values are consistent towards the root symbol.

\begin{algorithm}
    \caption{Expression hashing \cite{Burlacu2019eurocast}}\label{alg:expression-hashing}
    \SetKwInOut{Input}{input}\SetKwInOut{Output}{output}
    \SetKwFunction{Postorder}{Postorder}
    \SetKwFunction{Children}{Children}
    \SetKwFunction{Root}{Root}
    \SetKwFunction{Hash}{Hash}
    \SetKw{Continue}{continue}
    \SetKw{Return}{return}
    \Input{An expression $E$}
    \Output{The corresponding sequence of hash values}
    \BlankLine
    hashes $\gets$ empty list of hash values\;
    symbols $\gets$ list of symbols in $E$\;
    \ForEach{symbol $s$ in symbols}
    {
        $H(s) \gets$ an initial hash value\;
        \If{$s$ is a terminal function symbol}
        {
            \If{$s$ is commutative}{Sort the child nodes of $s$\;\label{alg:expression-hashing-sort}}
            child hashes $\gets$ hash values of $s$'s children\;
            $H(n) \gets \oplus{\left(\textup{child hashes}, H(s) \right)}$\;\label{alg:expression-hashing-compose-hash}
        }
        hashes.append($H(n)$)\;
    }
    \Return hashes\;
\end{algorithm}

\begin{figure}
    \centering
    \subfloat[Original expression]{\begin{tikzpicture}
    \tikzset{level distance=8mm}
    \Tree [.$+$ 
        [.$\times$ 
            [.$c$ ]
            \edge node[auto=left] { fold };
            [.$\times$
                [.$\exp$
                    [.$\times$ 
                        [.$c$ ]
                        [.$x_1$ ]
                    ]
                ]
                [.$\exp$
                    [.$\times$ 
                        [.$c$ ]
                        [.$x_1$ ]
                    ]
                ]
            ]
        ]
        \edge node[auto=left] { fold };
        [.$+$ 
            [.$c$ ]
            [.$\times$
                [.$c$ ]
                [.$x_1$ ]
            ]
            [.$\times$
                [.$c$ ]
                [.$x_1$ ]
            ]
        ]
    ]

\end{tikzpicture}}
    \subfloat[Folded expression]{\begin{tikzpicture}
    \tikzset{level distance=8mm}
    \Tree [.$+$ 
        [.$c$ ]
        [.$\times$
            [.$c$ ]
            [.$x_1$ ]
        ]
        [.$\times$
            [.$c$ ]
            [.$x_1$ ]
        ]
        [.$\times$ 
            [.$c$ ]
            [.$\exp$
            [.$\times$ 
                [.$c$ ]
                [.$x_1$ ]
            ]
            ]
            [.$\exp$
                [.$\times$ 
                    [.$c$ ]
                    [.$x_1$ ]
                ]
            ]
        ]
    ]

\end{tikzpicture}}
    \subfloat[Minimal form]{\begin{tikzpicture}
    \tikzset{level distance=8mm}
    \Tree [.$+$
        [.$c$ ]
        [.$\times$
            [.$c$ ]
            [.$x_1$ ]
        ]
        [.$\times$ 
            [.$c$ ]
                [.$\exp$
                [.$\times$ 
                    [.$c$ ]
                    [.$x_1$ ]
                ]
            ]
        ]
    ]
\end{tikzpicture}}
    \caption{Simplification to canonical minimal form during hashing}\label{fig:example-simplification}
\end{figure}
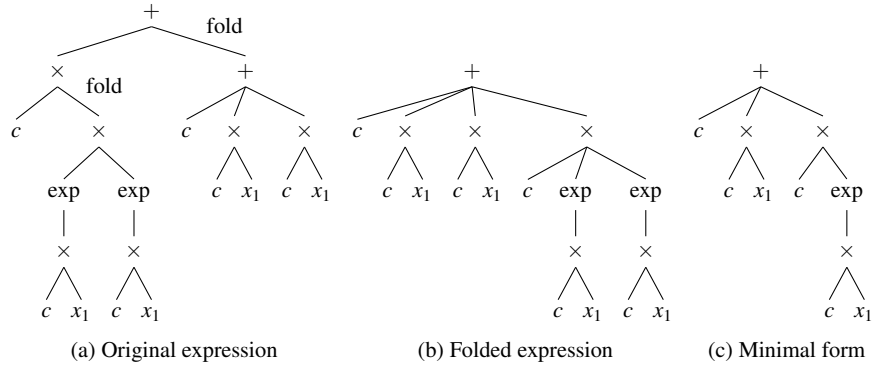

An expression's hash value is then given by the hash value of its root symbol. After sorting, sub-expressions with the same hash value are considered isomorphic and are simplified according to arithmetic rules. The simplification procedure is illustrated in Figure~\ref{fig:example-simplification} and consists of the following steps:

\begin{enumerate}
    \item \textbf{Fold}: Apply associativity to eliminate nested symbols of the same type. For example, postfix expression \texttt{a b + c +} consists of two nested additions where each addition symbol has arity 2. Folding flattens this expression to the equivalent form \texttt{a b c +} where the addition symbol has arity 3.
    \item \textbf{Simplify}: Apply arithmetic rules and mathematical identities to further simplify the expressions. Since expression already include placeholders for numerical coefficients, we eliminate redundant subexpressions such as \texttt{a a b +} which becomes \texttt{a b +}, or \texttt{a a +} which becomes \texttt{a}.
    \item Repeat steps 1 and 2 until no further simplification is possible.
\end{enumerate}

Nested $+$ and $\times$ symbols in Figure~\ref{fig:example-simplification} are folded in the first step, simplifying the tree structure of the expression. Arithmetic rules are then applied for further simplification. In this example, the product of exponentials 
\begin{equation*}
\exp(c_1 \times x_1) \times \exp(c_2 \times x_1) \equiv \exp((c_1+c_2) \times x_1)
\end{equation*} is simplified since from a local optimization perspective, optimizing the coefficients of the expression yields better results for a single coefficient $c_3 = c_1 + c_2$, thus it makes no sense to keep both original factors. Finally, the sum $c_4 x_1 + c_5 x_1$ is also simplified since one term in the sum is redundant.

After simplification, the hash value of the simplified tree is returned as the hash value of the original expression. Based on this computation we are able to identify already explored search paths and avoid duplicated effort.

\section{Exploring the Search Space}
\label{sec:exploring_search_space}





By limiting the size of expressions, the grammar and the hashing scheme produce a large but finite search space of semantically unique expressions. In an exhaustive search, we iterate all these expressions and search for the best fitting one. Thereby, we derive sentences with every possible derivation path. An expression is rejected if another expression with the same semantic---according to hashing---has already been generated during the search. When a new, previously unseen sentence is derived, the placeholders for coefficients are replaced with real values and optimized separately. The best fitting sentence is stored.

\begin{algorithm}
    \caption{Iterating the Search Space}
    \label{alg:search_space_search}
    
    \SetKwInOut{Input}{input}\SetKwInOut{Output}{output}
    \SetKw{Return}{return}

    \Input{Data set $ds$, max. number of variable references $\mathit{maxVariableRefs}$}
    \Output{Best fitting expression}
    \BlankLine
    $openPhrases \gets$ empty data structure\;
    $seenHashes \gets$ empty set\;
    Add $StartSymbol$ to $openPhrases$\;
    $bestExpression \gets$ constant symbol\; 
    \BlankLine
    \While{$openPhrases$ \textnormal{is not empty}}
    {
        $oldPhrase \gets$ fetch and remove from $openPhrases$\;
        $nonTerminalSymbol \gets$ leftmost nonterminal symbol in $oldPhrase$\;
        \ForEach{\textnormal{production} $prod$ \textnormal{of} $nonTerminalSymbol$}
        {
            $newPhrase \gets$ apply $prod$ on copy of $oldPhrase$\;
            \If{\textnormal{VariableRefs(}$newPhrase$\textnormal{)} $\le$ $\mathit{maxVariableRefs}$}{
                $hash \gets$ Hash($newPhrase$)\;
                \If{$seenHashes$ \textnormal{not contains} $hash$}{
                    Add $hash$ to $seenHashes$\;
                    \If{$newPhrase$ \textnormal{is sentence}}
                    {
                        Fit coefficients of $newPhrase$ to $ds$\;
                        Evaluate $newPhrase$ on $ds$\;
                        \If{$newPhrase$ \textnormal{is better than} $bestExpression$} {
                            $bestExpression \gets newPhrase$\;
                        }
                    }\Else{
                        Add $newPhrase$ to $openPhrases$\;
                    }
                }
            }
        }
    }
    \Return $bestExpression$
\end{algorithm}

Algorithm~\ref{alg:search_space_search} outlines how all unique expressions are derived: We store unfinished phrases---expressions with non-terminal symbols---in a data structure such as a stack or queue. We fetch phrases from this data structure one after another, derive new phrases, calculate their hash values and compare these hash values to previously seen ones. To derive new phrases, we always replace the \emph{leftmost} non-terminal symbol in the old phrase with the production rules of this non-terminal symbol. If a derived phrase becomes a sentence with only terminal symbols, its coefficients are optimized and its fitness is evaluated. Otherwise, if it still contains derivable non-terminal symbols, it is put back on the data structure. 

We restrict the length of a phrase by its number of variable references---e.g.~$x x$ and $\log(x) + x$ have two variable references. Phrases that exceed this limit are discarded in the search. Since every non-terminal symbol is eventually derived to at least one variable reference, non-terminal symbols count as variable references. In our experiments, a limit on the complexity has been found to be the most intuitive way to estimate an appropriate search space limit. Other measures, e.g.~the number of symbols are harder to estimate since coefficients, function symbols and the non-factorized representation of expression quickly inflate the number of symbols in a phrase.

\subsection{Symbolic Regression as Graph Search Problem}

\begin{figure}
    \centering
    \includegraphics[width=\textwidth]{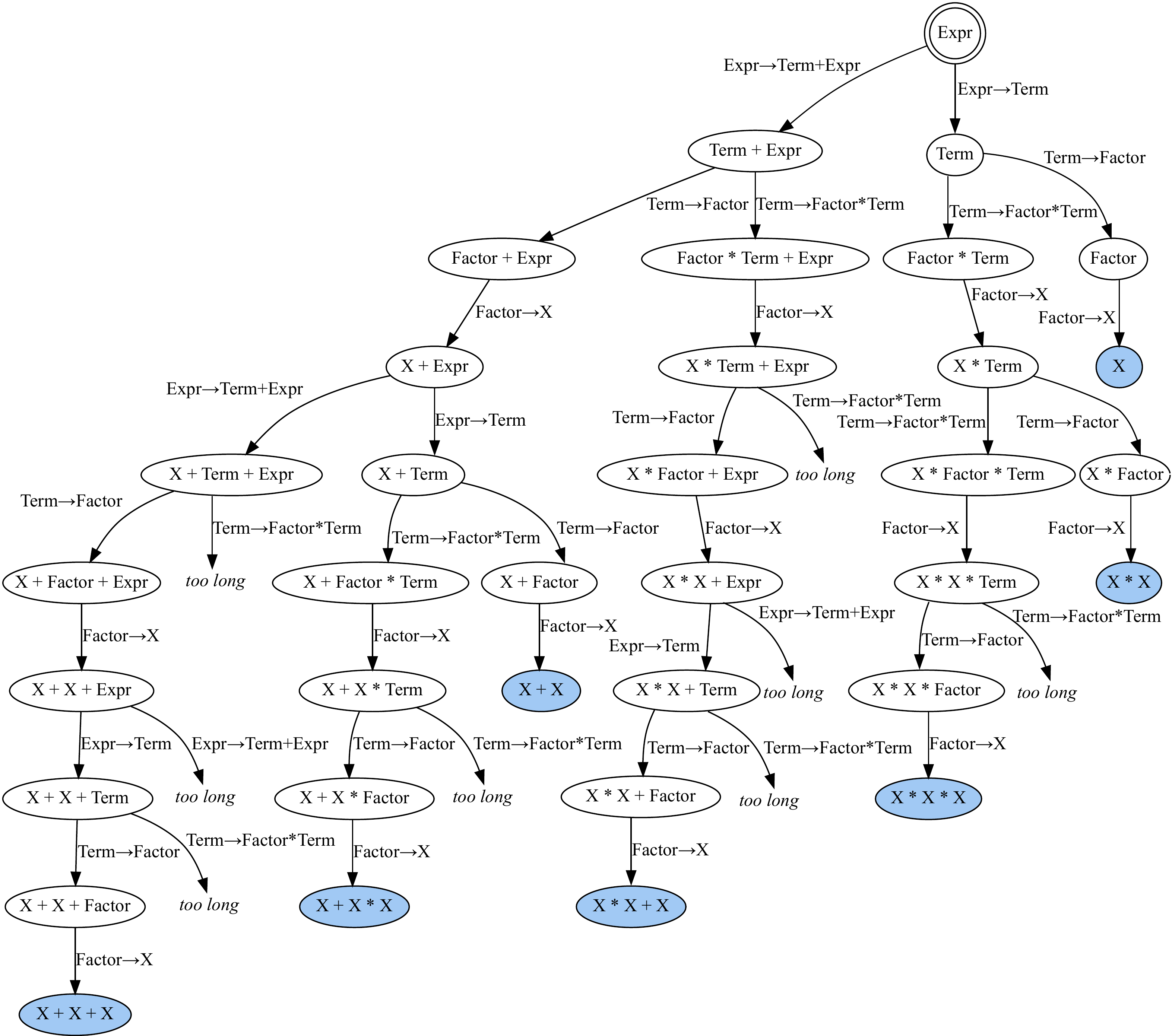}
    \caption{Search tree of expression generation without semantic hashing.}
    \label{fig:search_tree}
\end{figure}

\begin{figure}
    \centering
    \includegraphics[width=\textwidth]{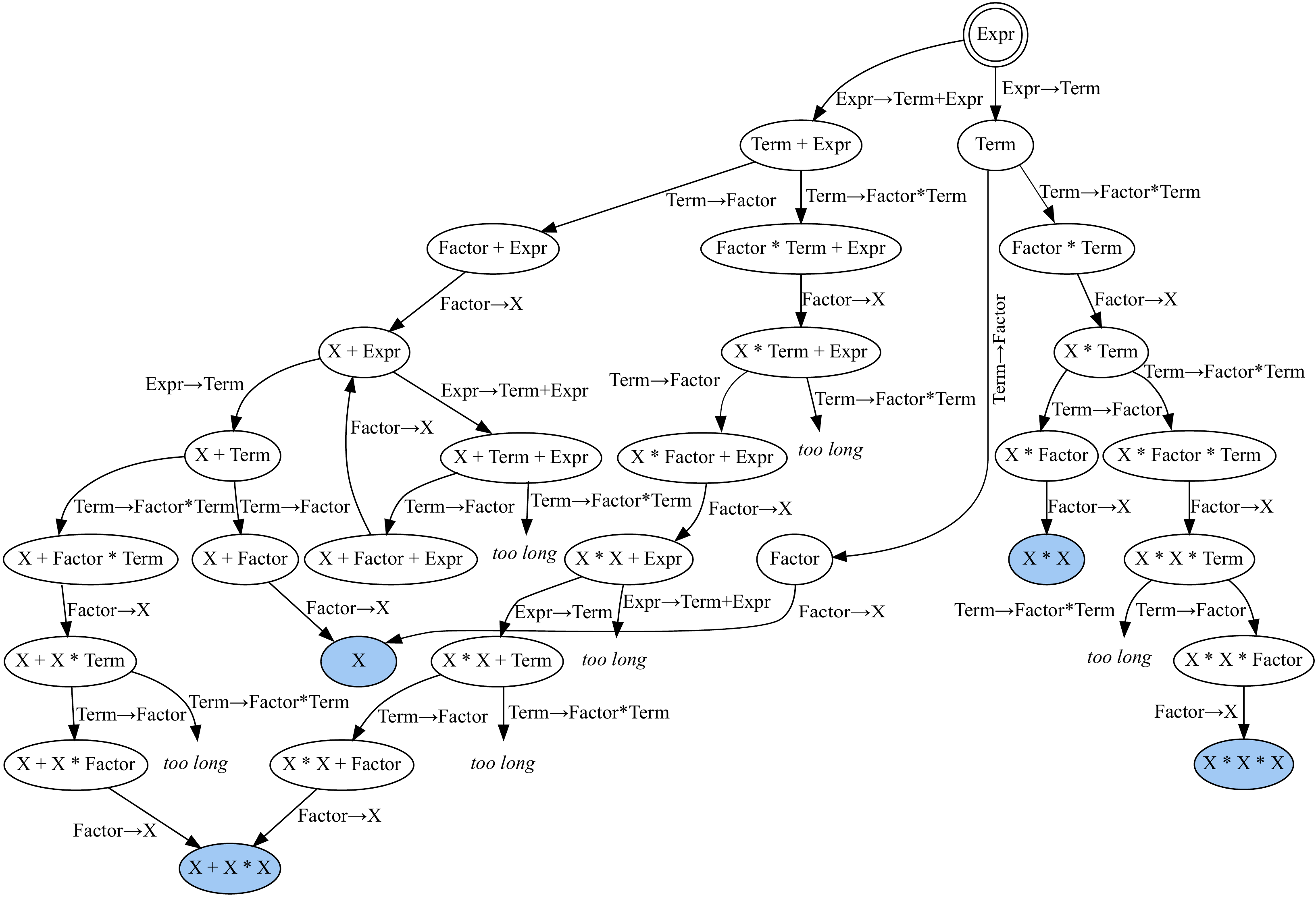}
    \caption{Search graph with loops caused by semantic hashing.}
    \label{fig:search_graph}
\end{figure}

Without considering the semantics of an expression, we would end up exploring a search tree like in Figure~\ref{fig:search_tree}, in which semantically equivalent expressions are derived multiple times (e.g.~$c_1x+c_2x$ and $c_1x+c_2x+c_3x$). However, hashing turns the search tree into a directed search graph in which nodes (derived phrases) are reachable via one or more paths, as shown in Figure~\ref{fig:search_graph}. Thus, hashing prevents the search in a graph region that was already visited. From this point of view, Algorithm~\ref{alg:search_space_search} is very similar to simple graph search algorithms such as depth-first or breadth-first search.

\subsection{Guiding the Search}

In Algorithm~\ref{alg:search_space_search}, the order in which expressions are generated is determined by the data structure used. A stack or a queue would result in a depth-first or a breadth-first search respectively. However, as the goal is to find well-fitting expressions quickly and efficiently, we need to guide the traversal of a search graph towards promising phrases.

Our general framework for guiding the search is very similar to the idea used in the A* algorithm~\cite{hart1968formal}. We use a priority queue as data structure and assign a priority value to each phrase, indicating the expected quality of sentences which are derivable from that phrase. Phrases with high priority are derived first in order to discover well-fitting sentences, steering the algorithm towards good solutions.

Similar to the A* algorithm, we cannot make a definite statement about a phrase's priority before actually deriving all possible sentences from it. Therefore, we need to estimate this value with problem-specific heuristics. The calculation of phrase priorities provides us a generic point for integrating heuristics for improving the search efficiency and extending the algorithm's capabilities in future work.

\section{Steering the Search}
\label{sec:search_heuristic}


We introduce a simple heuristic for guiding the search and leave more complex and efficient heuristics for future work. The proposed heuristic makes a pessimistic estimation of the quality of a phrase's derivable sentences. This is done by evaluating phrases before they are derived to sentences. With the goal of finding short and accurate sentences quickly, the priority value considers both the expected quality and the length of a phrase. 

\subsection{Quality Estimation}

Estimating the expected quality of an unfinished phrase is possible due to the polynomial structure of sentences and the derivation of the leftmost non-terminal symbol in every phrase. Since expressions are sums of terms ($c_1 \mathit{Term}_1 + c_2 \mathit{Term}_2 + ...$), repeated expansion of the leftmost non-terminal symbol derives one term after another. This results in phrases such as in Equation~\ref{eq:unfinished_phrase}, in which the first two terms $c_1 \log(c_2 x + c_3)$ and $c_4 x x$ contain only terminal symbols and the last non-terminal symbol is \textit{Expr}.

\begin{equation}
    \stackrel{\mathit{finished Term}_1}{c_1 \log(c_2 x + c_3)} 
    \quad + \quad
    \stackrel{\mathit{finished Term}_2}{c_4 x x\vphantom{\log()}}
    \quad + 
    \underbrace{
        \mathit{Expr}
    }_{\text{Treat as coefficient}}
    \label{eq:unfinished_phrase}
\end{equation}

Phrases where the only non-terminal symbol is \texttt{Expr} are evaluated as if they were full sentences by treating \texttt{Expr} as a coefficient during the local optimization phase.
We get a pessimistic estimate of the quality of derivable sentences, since derived sentences with more terms can only have better quality. The quality can only improve with more terms because of separate coefficient optimization and one scaling coefficient per term, as shown in Equation~\ref{eq:unfinished_phrase_abstract}. If a term which does not improve the quality is derived, the optimization of coefficients will cancel it out by setting the corresponding scaling coefficient to zero (e.g.~$c_5$ in Equation~\ref{eq:unfinished_phrase_abstract}). 

\begin{equation}
    \mathit{finished Term}_1 
    \quad + \quad
    \mathit{finished Term}_2
    \quad + 
    \underbrace{c_5 \mathit{Term}}_{\text{Can only improve quality}
    }
    \label{eq:unfinished_phrase_abstract}
\end{equation}

This heuristic works only for phrases in which \textit{Expr} is the only non-terminal symbol. For sentences with different non-terminal symbols, we reuse the estimated quality from the last evaluated parent phrase. The estimate is updated when a new term with only terminal symbols is derived and again only one \textit{Expr} remains. For now, we do not have a reliable estimation method for terms that contain non-terminal symbols and leave this topic for future work.

\subsection{Priority Calculation}

To prevent arbitrary adding of badly-fitting terms that are eventually scaled down to zero, our priority measure considers both a phrase's length and its expected accuracy. To balance these two factors, these two measures need to be in the same scale. We use the normalized mean squared error (NMSE) as quality measure which is in the range $[0,1]$ for properly scaled solutions. This measure corresponds to $1 - R^2$ (coefficient of determination). As length measure we use the number of symbols relative to the maximum sentence length.

Since we limit the search space to a maximum number of variable references of a phrase, we cannot exactly calculate the maximum possible length of a phrase. Therefore, we estimate this maximum length with a greedy procedure: Starting with the grammar's start symbol \textit{Expr}, we iteratively derive a new phrase using the longest production rule. If two production rules have the same length, we take the one with least non-terminal symbols and variable references.

Phrases with \emph{lower} priority values are expanded first during the search. The priority for steering the search from Section~\ref{sec:exploring_search_space} is the phrase's $\operatorname{NMSE}$ value minus its weighted relative length, as shown in Equation~\ref{eq:priority}. The weight $w$ controls the greediness and allows corrections of over- or underestimations of the maximum length. However, in practice this value is not critical.

\begin{equation}
\label{eq:priority}
\operatorname{priority(\mathit{p})} \ = \ \operatorname{NMSE}(p) \ - \ w \frac{\operatorname{len}(\mathit{p})}{\mathit{length}_{\max}}
\end{equation}

\section{Experiments}
\label{sec:experiments}

We run our algorithm on several synthetic benchmark datasets to show that the search space defined by our restricted grammar is powerful enough to solve many problems in feasible time. As benchmark datasets, we use noiseless datasets from physical domains \cite{CHEN:2018:ESA} and Nguyen-, Vladislavleva- and Keijzer-datasets \cite{White2013} as defined and implemented in the HeuristicLab framework.

The search space was restricted in the experiments to include only sentences with at most 20 variable references. We evaluate at most 200~000 sentences. Coefficients are randomly initialized and then fitted with the iterative gradient-based Levenberg-Marquardt algorithm \cite{levenberg1944method, marquardt1963algorithm} with at most 100 iterations. For each model structure, we repeat the coefficient fitting process ten times with differently initialized values to reduce the chance of finding bad local optima.

As a baseline, we also run symbolic regression with GP on the same benchmark problems. Therefore, we execute GP with strict offspring selection (OSGP) \cite{affenzeller:2009} and explicit optimization of coefficients \cite{Kommenda:2013}. The OSGP settings are listed in Table \ref{tab:osgp_settings}.  The OSGP experiments were executed with the \textit{HeuristicLab} software framework\footnote{\url{https://dev.heuristiclab.com}} \cite{wagner2005heuristiclab}. Since this comparison focuses only on particular weaknesses and strengths of our proposed algorithm over state of the art-techniques, we use the same OSGP settings for all experiments and leave out problem-specific hyper parameter-tuning. 

\begin{table}[t]
    \caption{OSGP experiment settings}
    \label{tab:osgp_settings}
    \begin{tabular}{p{.35\textwidth}p{.63\textwidth}}
        \hline\noalign{\smallskip}
        Parameter & Setting \\
        \noalign{\smallskip}\svhline\noalign{\smallskip}
        Population size         & 500 \\
        Max.~selection pressure & 300 \\
        Max.~evaluated solutions & 200 000 \\
        Mutation probability    & 15\% \\
        Selection               & Gender-specific selection (random and proportional) \\
        Crossover operator      & Subtree swapping \\
        Mutation operator       & Point mutation, tree shaking, changing single symbols, replacing/removing branches \\
        Max.~tree size          & Number of nodes: 30, depth: 50 \\
        Function set       & $+, -, \times, \div $, exp, log, sin, cos, square, sqrt, cbrt \\
        \noalign{\smallskip}\hline\noalign{\smallskip}
    \end{tabular}    
\end{table}

\subsection{Results}

Both the exhaustive search and OSGP were repeated ten times on each dataset. All repetitions of the exhaustive search algorithm led to the exact same results. This underlines the determinism of the proposed methods, even though we rely on stochasticity when optimizing coefficients. Also the OSGP results do not differ much. Tables \ref{tab:keijzer}-\ref{tab:other_benchmarks} show the achieved NMSE values for the exhaustive search and the median NMSE values of all OSGP repetitions. NMSE values in the Tables \ref{tab:keijzer}-\ref{tab:other_benchmarks} smaller than $10^{-8}$ are considered as exact or good-enough approximations and emphasized in bold. The exhaustive search found a good solution (NMSE $< 10^{-8}$) within ten minutes for all datasets. If no such solution was found, the algorithm runs until it reaches the max.~number of evaluated solutions, which can take days for larger datasets.

\begin{table}[t] 
    \caption{Median NMSE results for Keijzer instances.}
    \label{tab:keijzer}
    \begin{tabular}{p{1.3cm}p{5.6cm}p{1cm}p{1.4cm}p{1cm}p{.7cm}}
\hline\noalign{\smallskip}
{} &                                                             Problem & \multicolumn{2}{l}{Exhaustive Search} & \multicolumn{2}{c}{OSGP} \\
{} & & Train. &            Test & Train. &            Test \\
\noalign{\smallskip}\svhline\noalign{\smallskip}
 1 \cite{keijzer03,keijzer:2000:GPbvt} &  $0.3  x \sin(2  \pi  x); x \in [-1, 1]$ &  3e-27 &  \textbf{2e-27} &  1e-30 &  \textbf{8e-31} \\
 2 \cite{keijzer03} &  $0.3  x \sin(2  \pi  x); x \in [-2, 2]$ &  5e-22 &  \textbf{5e-22} &  5e-18 &  \textbf{4e-18} \\
 3 \cite{keijzer03} &  $0.3  x \sin(2  \pi  x); x \in [-3, 3]$ &  6e-32 &  \textbf{3e-31} &  4e-30 &  \textbf{3e-30} \\
 4 \cite{keijzer03,Salustowicz:97ecj} &  $x^3   \exp(-x)  \cos(x)  \sin(x)  (\sin(x)^2  \cos(x) - 1)$ &  1e-04 &  2e-04 &  1e-06 &  1e-06 \\
 5 \cite{keijzer03} &  $(30  x  z) / ((x - 10)   y^2)$ &  3e-08 &  3e-08 &  3e-20 &  \textbf{3e-20} \\
 6 \cite{keijzer03,streeter:masters} &  $\sum_{i=1}^x \frac{1}{i}$ &  8e-13 &  \textbf{6e-09} &  5e-14 &  \textbf{5e-13} \\
 7 \cite{keijzer03,streeter:masters} &  $\ln(x)$ &  2e-31 &  \textbf{3e-31} &  1e-30 &  \textbf{2e-30} \\
 8 \cite{keijzer03,streeter:masters} &  $\sqrt{x}$ &  2e-14 &  \textbf{8e-10} &  5e-21 &  \textbf{1e-21} \\
 9 \cite{keijzer03,streeter:masters} &  $\operatorname{arcsinh}(x)  \ \text{i.e.} \ln(x + \sqrt{x^2 + 1})$ &  5e-14 &  1e-05 &  5e-17 &  \textbf{6e-16} \\
 10 \cite{keijzer03,streeter:masters} &  $x ^ y$ &  4e-04 &  1e-01 &  6e-32 &  2e-04 \\
 11 \cite{keijzer03,topchy:2001:gecco} &  $xy + \sin((x - 1)(y - 1))$ &  7e-04 &  7e-01 &  2e-22 &  9e-02 \\
 12 \cite{keijzer03,topchy:2001:gecco} &  $x^4 - x^3 + y^2 / 2 - y$ &  5e-32 &  \textbf{1e-31} &  7e-22 &  \textbf{8e-18} \\
 13 \cite{keijzer03,topchy:2001:gecco} &  $6  \sin(x)  \cos(y)$ &  2e-32 &  \textbf{2e-31} &  3e-32 &  \textbf{3e-32} \\
 14 \cite{keijzer03,topchy:2001:gecco} &  $8 / (2 + x^2 + y^2)$ &  4e-32 &  \textbf{2e-31} &  1e-17 &  \textbf{1e-17} \\
 15 \cite{keijzer03,topchy:2001:gecco} &  $x^3 / 5 + y^3 / 2 - y - x$ &  1e-22 &  \textbf{2e-21} &  2e-11 &  \textbf{6e-10} \\
\noalign{\smallskip}\hline\noalign{\smallskip}
\end{tabular}
  
\end{table}

The experimental results show, that our algorithm struggles with problems with complex terms---for example with Keijzer data sets 4, 5 and 11 in Table \ref{tab:keijzer}. This is probably because our heuristic works ''term-wise''---our algorithm searches completely broad without any guidance within terms which still contain non-terminal symbols. This issue becomes even more pronounced when we have to find long and complex function arguments. It should also be noted that our algorithm only finds non-factorized representations of such arguments, which are even longer and therefore even harder to find in a broad search.

\begin{table}[b]
    \caption{Median NMSE results for Nguyen instances.}
    \label{tab:nguyen}
    \begin{tabular}{p{0.9cm}p{6cm}p{1cm}p{1.4cm}p{1cm}p{.7cm}}
\hline\noalign{\smallskip}
{} &                            Problem & \multicolumn{2}{l}{Exhaustive Search} & \multicolumn{2}{c}{OSGP} \\
{} & & Train. &            Test & Train. &            Test \\
\noalign{\smallskip}\svhline\noalign{\smallskip}
 1 \cite{Quang:2011:GPEM} &  $x^3 + x^2 + x$ &  5e-34 &  \textbf{3e-33} &  8e-30 &  \textbf{2e-29} \\
 2 \cite{Quang:2011:GPEM} &  $x^4 + x^3 + x^2 + x$ &  3e-33 &  \textbf{4e-33} &  5e-30 &  \textbf{1e-28} \\
 3 \cite{Quang:2011:GPEM} &  $x^5 + x^4 + x^3 + x^2 + x$ &  1e-33 &  \textbf{7e-33} &  2e-16 &  \textbf{2e-15} \\
 4 \cite{Quang:2011:GPEM} &  $x^6 + x^5 + x^4 + x^3 + x^2 + x$ &  6e-12 &  \textbf{6e-11} &  2e-12 &  3e-08 \\
 5 \cite{Quang:2011:GPEM} &  $\sin(x^2)\cos(x) - 1$ &  9e-14 &  \textbf{3e-13} &  3e-18 &  \textbf{4e-18} \\
 6 \cite{Quang:2011:GPEM} &  $\sin(x) + \sin(x + x^2)$ &  2e-17 &  \textbf{2e-12} &  6e-14 &  6e-08 \\
 7 \cite{Quang:2011:GPEM} &  $\log(x + 1) + \log(x^2 + 1)$ &  4e-13 &  \textbf{5e-12} &  5e-13 &  \textbf{1e-09} \\
 8 \cite{Quang:2011:GPEM} &  $\sqrt{x}$ &  6e-32 &  \textbf{2e-31} &  7e-32 &  \textbf{1e-31} \\
 9 \cite{Quang:2011:GPEM} &  $\sin(x) + \sin(y^2)$ &  2e-13 &  \textbf{2e-12} &  8e-31 &  \textbf{8e-31} \\
 10 \cite{Quang:2011:GPEM} &  $2\sin(x)\cos(y)$ &  5e-32 &  \textbf{1e-31} &  1e-28 &  \textbf{8e-29} \\
 11 \cite{Quang:2011:GPEM} &  $x^y$ &  2e-06 &  1e-02 &  6e-30 &  \textbf{3e-30} \\
 12 \cite{Quang:2011:GPEM} &  $x^4 - x^3 + y^2/2 - y$ &  2e-31 &  \textbf{2e-31} &  7e-18 &  \textbf{5e-17} \\
\noalign{\smallskip}\hline\noalign{\smallskip}
\end{tabular}
  
\end{table}

For the Nguyen datasets in Table \ref{tab:nguyen} and the Keijzer datasets 12-15 in Table \ref{tab:keijzer}, we find the exact or good approximations in most cases with our exhaustive search. Especially for simpler datasets, the results of our algorithm surpasses the one of OSGP. This is likely due to the datasets' low number of training instances, which makes it harder for OSGP to find good approximations.

\begin{table}[t]
    \caption{Median NMSE results for Vladislavleva instances.}
    \label{tab:vladislavleva}
    \begin{tabular}{p{0.9cm}p{6cm}p{1cm}p{1.4cm}p{1cm}p{.7cm}}
\hline\noalign{\smallskip}
{} &                                                        Problem & \multicolumn{2}{l}{Exhaustive Search} & \multicolumn{2}{c}{OSGP} \\
{} & & Train. &            Test & Train. &            Test \\
\noalign{\smallskip}\svhline\noalign{\smallskip}
 1 \cite{smits:2004:GPTP} &  $\exp(- (x_1 - 1)^2) / (1.2 + (x_2 - 2.5)^2)$ &  3e-03 &  3e-01 &  1e-09 &  9e-07 \\
 2 \cite{Salustowicz:97ecj} &  $\exp(-x)  x^3  \cos(x)  \sin(x)  (\cos(x)\sin(x)^2 - 1)$ &  3e-04 &  1e-02 &  3e-06 &  2e-03 \\
 3 \cite{Vladislavleva:2009:TEC} &  $f_2(x_1)(x_2 - 5)$ &  1e-02 &  2e-01 &  3e-05 &  6e-04 \\
 4 \cite{Vladislavleva:2009:TEC} &  $10/(5 + \sum_{i=1}^5 (x_i - 3)^2)$ &  1e-01 &  2e-01 &  7e-03 &  1e-02 \\
 5 \cite{Vladislavleva:2009:TEC} &  $30  ((x_1 - 1)  (x_3 -1)) / (x_2^2  (x_1 - 10))$ &  2e-03 &  9e-03 &  8e-16 &  \textbf{9e-15} \\
 6 \cite{Vladislavleva:2009:TEC} &  $6  \sin(x_1)  \cos(x_2)$ &  8e-32 &  \textbf{4e-31} &  6e-31 &  \textbf{3e-19} \\
 7 \cite{Vladislavleva:2009:TEC} &  $(x_1 - 3)(x_2 - 3) + 2  \sin((x_1 - 4)(x_2 - 4))$ &  1e-30 &  \textbf{9e-31} &  5e-29 &  \textbf{4e-29} \\
 8 \cite{Vladislavleva:2009:TEC} &  $((x_1 - 3)^4 + (x_2 - 3)^3 - (x_2 -3)) / ((x_2 - 2)^4 + 10)$ &  1e-03 &  2e-01 &  5e-05 &  2e-02 \\
\noalign{\smallskip}\hline\noalign{\smallskip}
\end{tabular}
  
\end{table}

Some problems are not contained in the search space, thus we do not find any good solution for them. This is the case for Keijzer 6, 9 and 10 in Table \ref{tab:keijzer}, for which we do not support the required function symbols in our grammar. Also all Vladislavleva datasets except 6 and 7 in Table \ref{tab:vladislavleva} and the problems ''Fluid Flow'' and ''Pagie-1'' in Table \ref{tab:other_benchmarks} are not in the hypothesis space as they are too complex.

\begin{table}
    \caption{Median NMSE results for other instances.}
    \label{tab:other_benchmarks}
    \begin{tabular}{p{6.9cm}p{1cm}p{1.5cm}p{1cm}p{.7cm}}
\hline\noalign{\smallskip}
                                                                                                                       Problem & \multicolumn{2}{l}{Exhaustive Search} & \multicolumn{2}{c}{OSGP} \\
                                                                                                                               &            Train. &            Test & Train. &            Test \\
\noalign{\smallskip}\svhline\noalign{\smallskip}
 Poly-10 \cite{poli03} \newline $x_1 x_2 + x_3 x_4 + x_5 x_6 + x_1 x_7 x_9 + x_3 x_6 x_{10}$ &  2e-32 &  \textbf{1e-32} &  7e-02 &  1e-01 \\ \noalign{\smallskip}
 Pagie-1 (Inverse Dynamics) \cite{pagie97evolutionary} \newline $1/(1 + x^{-4}) + 1/(1 + y^{-4})$ &  1e-03 &  6e-01 &  9e-07 &  5e-05 \\ \noalign{\smallskip}
 Aircraft Lift Coefficient \cite{CHEN:2018:ESA}\newline $C_{L\alpha} (\alpha - \alpha_0) + C_{L \delta_e} \delta_e S_{HT} / S_{\mathit{ref}}$ &  3e-31 &  \textbf{3e-31} &  2e-17 &  \textbf{2e-17} \\ \noalign{\smallskip}
 Fluid Flow \cite{CHEN:2018:ESA}\newline $V_\infty r \sin(\theta) (1 - R^2/r^2) + \Gamma/(2 \pi) \ln(r/R)$ &  3e-04 &  4e-04 &  9e-06 &  2e-05 \\ \noalign{\smallskip}
 Rocket Fuel Flow \cite{CHEN:2018:ESA}\newline $ p_0 A^{\star} / \sqrt{T_0}  \sqrt{\gamma/R (2/(\gamma+1))^{(\gamma+1) / (\gamma-1)}}$ &  3e-31 &  \textbf{3e-31} &  1e-19 &  \textbf{1e-19} \\ 
\noalign{\smallskip}\hline\noalign{\smallskip}
\end{tabular}
  
\end{table}

Another issue is the optimization of coefficients. Although several problems have a simple structure and are in the search space, we do not find the right coefficients for arguments of non-linear functions, for example in Nguyen 5-7. The issue hereby is that we iterate over the actually searched model structure but determine bad coefficients. As we do never look again at the same model structure, we can only find an approximation. This is a big difference to symbolic regression with genetic programming, in which we might find the same structure again in next generations.

\section{Discussion}
\label{sec:discussion}

Among the nonlinear system identification techniques, symbolic
regression is characterized by its ability to identify complex
nonlinear relationships in structured numerical data in the form of
interpretable models. The combination of the power of nonlinear
system identification without a priori assumptions about the model
structure with the white-box ability of mathematical formulas
represents the unique selling point of symbolic regression. If tree-based
GP is used as search method,
the ability to interpret the found models is limited due to the
stochasticity of the GP search. Thus, at the end of the modeling
phase, several similarly complex models of approximately the same
quality can be produced, which have completely different structures
and use completely different subsets of features.  These
last-mentioned limitations due to ambiguity can be countered using a
deterministic approach in which only semantically unique models may be
used. This approach, however, requires a lot of restrictions regarding
search space complexity in order to specify a subspace in which an
exhaustive search is feasible. On the other hand, the exhaustive claim
enables the approach to generate extensive model libraries
already in the offline phase, through which as soon as a concrete task
is given in the online phase, it is only necessary to navigate in a
suitable way.

In a very reduced summary, one could characterize the classical tree-based
symbolic regression using GP and the approach of
deterministically and exhaustively generating models in such a way that the
latter enables a complete search in an incomplete search space
while the classical approach performs an incomplete search in a rather
complete search space.


\subsection{Limitations}

The approach we have described in this contribution also has several
limitations. For the identification of optimal coefficient values we rely on the
Levenberg-Marquardt method for least squares, which is a local search
routine using gradient information. Therefore, we can only hope to
find global optima for coefficient values. Finding bad local optima for coefficients is
less of a concern when using GP variants with a similar local
improvement scheme because there are implicitly many restarts through
the evolutionary operations of recombination and mutation. In the
proposed method we visit each structure only once and therefore risk
to discard a good solution when we are unlucky to find good coefficients.

We have worked only with noiseless problem instances yet. We observed in first experiments with noisy problems instances that the algorithm might get stuck trying to improve non-optimal partial solutions due to its greedy nature. Therefore, we need further investigations before we move on with the development of our algorithm to noisy real-world problems.

Another limitation is the poor scalability of grammar enumeration
when increasing the number of features or the size of the search
space. When increasing these parameters we can not expect to explore a
significant part of the complete search space and must increasingly
rely on the power of heuristics to hone in on relevant subspaces.
Currently, we have only integrated a single heuristic which evaluates
terms in partial solutions and prioritizes phrase which include
well-fitting terms. However, the algorithm has no way to prioritize incomplete terms and is inefficient when trying to find complex terms.

\section{Outlook} 

%

Even when considering the above mentioned limitations of the currently
implemented algorithm we still see significant potential in the
approach of more systematic and deterministic search for symbolic
regression and we already have several ideas to improve the algorithm
and overcome some of the limitations.

The integration of improved heuristics for guided search is our
top-priority. An advantage of the concept is that it is extremely general
and allows to experiment with many different heuristics. Heuristics
can be as simple as prioritizing shorter expressions or less complex
expressions. More elaborate schemes which guide the search based on
prior knowledge about the data-generating process are easy to
imagine. Heuristics could incorporate syntactical information
(e.g. which variables already occur within the expression) as well as
information from partial evaluation of expressions. We also consider
dynamic heuristics which are adjusted while the algorithm is running
and learning about the problem domain.
Potentially, we could even identify and learn heuristics
which are transferable to other problem instances and would improve
efficiency in a transfer learning setting.

Getting trapped in local optima is less of a concern when we apply
global search algorithms for coefficient values such as evolution
strategies, differential evolution, or particle swarm optimization
(cf. \cite{Korns:2010:GPTP}). Another approach would be to reduce the
ruggedness of the objective function through regularization of the
coefficient optimization step. This could be helpful to reduce the
potential of overfitting and getting stuck in sub-optimal subspaces
of the search space.

Generally, we consider grammar enumeration to be effective only when
we limit the search space to relatively short expressions---which is often the
case in our industrial applications. Therein lies the main potential
compared to the more general approach of genetic programming. In this
context we continue to explore potential for segmentation of the
search space \cite{Kronberger2019} in combination with grammar
enumeration in an offline phase for improving later search
runs. Grammar enumeration with deduplication of structures could also
be helpful to build large offline libraries of sub-expressions that
could be used by GP
\cite{angeline:1993:ema,eurogp:KeijzerRMC05,Krawiec:2012:GECCOcomp,chrisgptp2015behavioral}.

\begin{acknowledgement}
    
The authors gratefully acknowledge support by the Christian Doppler Research Association and the Federal Ministry for Digital and Economic Affairs within the \emph{Josef Ressel Center for Symbolic Regression}.
\end{acknowledgement}


\end{document}